\documentclass[letterpaper, 10 pt, conference]{ieeeconf}  % Comment this line out if you need a4paper
\IEEEoverridecommandlockouts                              % This command is only needed if 
\usepackage{url}
\usepackage{graphics} % for pdf, bitmapped graphics files
\usepackage{epsfig} % for postscript graphics files
\usepackage{mathptmx} % assumes new font selection scheme installed
\usepackage{amsmath} % assumes amsmath package installed
\usepackage{cite}
\usepackage{siunitx}
\usepackage{soul}
\usepackage{color}
\usepackage{xcolor}

\soulregister\ref7  % 针对\ref命令
\soulregister\cite7 
\soulregister\eqref7
\soulregister\prettyref7

\usepackage{algorithm} 
\usepackage{algpseudocode}
\usepackage{makecell} % for word position in the table
\usepackage{multirow} % table
\usepackage{booktabs} % table
\usepackage{threeparttable} % table
\usepackage{diagbox}  % diagonal line in table

\bstctlcite{IEEEexample:BSTcontrol}

\usepackage{prettyref}
\newrefformat{fig}{Fig.~\ref{#1}}
\newrefformat{sec}{Sec.~\ref{#1}}
\newrefformat{tab}{Tab.~\ref{#1}}
\newrefformat{algo}{Algo.~\ref{#1}}
\newrefformat{eq}{(\ref{#1})}
\newrefformat{app}{\ref{#1}} % BZZ

\title{\LARGE \bf
CafkNet: GNN-Empowered Forward Kinematic Modeling for Cable-Driven Parallel Robots}
\author{Zeqing Zhang$^{1,\dagger}$, Linhan Yang$^{1,\dagger}$, Cong Sun$^2$, Weiwei Shang$^3$ and Jia Pan$^1$%
\thanks{$^1$ The University of Hong Kong, Hong Kong, China.}%
\thanks{$^2$ Northeastern University, Shenyang, China.}
\thanks{$^3$ University of Science and Technology of China, Hefei, China.}
\thanks{$^\dagger$ These authors contributed equally to this work.}
\thanks{Corresponding author: Jia Pan {\tt\small jpan@cs.hku.hk}}
}

\begin{document}
\maketitle
\thispagestyle{empty}
\pagestyle{empty}

%%%%%%%%%%%%%%%%%%%%%%%%%%%%%%%%%%%%%%%%%%%%%%[section]%%%%%%%%%%%%%%%%%%%%%%%%%%%%%%%%%%%%%%%%%%%%%%%%%%%%%%%%%%
\begin{abstract}
Cable-driven parallel robots (CDPRs) have gained significant attention due to their promising advantages. When deploying CDPRs in practice, the kinematic modeling is a key question. Unlike serial robots, CDPRs have a simple inverse kinematics problem but a complex forward kinematics (FK) issue. So, the development of accurate and efficient FK solvers has been a prominent research focus in CDPR applications. By observing the topology within CDPRs, in this paper, we propose a graph-based representation to model CDPRs and introduce CafkNet, a fast and general FK solving method, leveraging Graph Neural Network (GNN) to learn the topological structure and yield the real FK solutions with superior generality, high accuracy, and low time cost.
CafkNet is extensively tested on 3D and 2D CDPRs in different configurations, both in simulators and real scenarios. The results demonstrate its ability to learn CDPRs' internal topology and accurately solve the FK problem. Then, the zero-shot generalization from one configuration to another is validated. Also, the sim2real gap can be bridged by CafkNet using both simulation and real-world data.
To the best of our knowledge, it is the first study that employs the GNN to solve FK problem for CDPRs.
Videos and codes are available at \url{https://sites.google.com/view/cafknet/site}.
\end{abstract}

\section{Introduction}
\label{sec:intro}

Cable-driven parallel robots (CDPR) is a kind of parallel mechanism that employs cables to replace rigid links \cite{qian2018review,zhang2022state,zarebidoki2022review,zhang2024synthetic}. It has exhibited excellent scalability in terms of size, payload, and dynamics capacities. 

The crucial step before deploying CDPRs is modeling, including kinematic modeling (KM) and dynamics modeling (DM). KM refers to the process of creating a mathematical equation of the end-effector movement, including its kinematic parameters such as position, orientation, and velocity. DM, on the other hand, is the process of creating a mathematical representation of how the end-effector will respond to external forces over time. For the KM of CDPR, the usual practice, like \cite{pott2008forward}, is to first determine the configuration of the CDPR according to the application requirements, including the number of cables, positions of the pulley on the framework, the shape of the end-effector and the position on which the cable is connected. Then, according to the geometric relationship, the modeling of cable attachment on the end-effector and anchor points on the framework is formulated. After that, some downstream tasks can be completed, such as workspace determination \cite{pham2006force}, path planning \cite{zhang2019efficient}, and control \cite{zhang2023design}.
In CDPR's KM, there are typically two types of calculations: forward kinematics (FK) and inverse kinematics (IK). FK involves determining the end-effector pose (i.e., position and orientation) based on cable lengths, while IK solves for the length of cables given the end-effector pose in reverse. In contrast to the traditional serial mechanical structure, the FK problem of CDPRs is more complex than its IK.

Therefore, numerous works focus on solving FK problems of CDPRs. \cite{pott2015forward} discusses the non-trivial computation of the FK of CDPRs, which are kinematically over-constrained mechanisms. In addition, \cite{qian2018review} provides the overview of FK solvers based on successful application cases of CDPRs. Recent work \cite{mishra2022forward} proposes a neural network (NN) approach for solving the FK of under-constrained CDPRs. In summary, solving the FK for CDPRs presents several challenges.
\begin{enumerate}
    \item Nonlinearity. Solving the FK problem for CDPRs, especially when over-constrained, is challenging due to its non-linear nature in FK formulations.
    % The difficulty in solving the FK problem for CDPRs, especially for kinematically over-constrained ones, lies in its complex and non-linear nature, making it challenging to find a solution using traditional analytical or numerical methods from kinematic equations.
    \item Flexibility of Configuration. CDPRs offer customizable cable configurations, but modeling each configuration separately creates additional work.
    % The flexibility of CDPR is that its configuration (i.e., positions of cables attached to the framework and end-effector) can be customized according to application requirements. But it also brings a lot of work to KM, since each configuration needs to be modeled separately. There is a lack of a generalized FK approach.
    \item Sim2Real Gap. The gap between the simulated and real-world performance of CDPRs is inevitable due to limited model accuracy and complex cable systems.
    % It refers to the gap between the simulation results and the real-world performance of a physical CDPR system. The challenges in Sim2Real include limited accuracy of simulation models and difficulty in accurately modeling complex cable routing, pulley, and winch systems in CDPRs.
\end{enumerate}
Due to the nonlinearity and complexity of FK problems, FK solvers have a large computational burden and also suffer from the problem of no solution or multiple solutions. But in practical applications, people may focus on the real solution determination and not ask for all solutions, which, however, is the main purpose of existing FK solvers. Furthermore, configuration-dependent FK solvers may not be easily applied to the reconfiguration CDPR \cite{wang2023optimal} or the case of adding or subtracting cables. However, according to our observations in \cite{zhang2022ray}, we notice that the change in CDPR's configuration does not affect the topological architecture if its configuration can be represented as a graph. In addition, it is an indisputable fact that the Sim2Real gap is inevitable, despite significant efforts to reduce differences \cite{liu2023dynamic,akhmetzyanov2022model,wang2022real2sim2real}. This is due to the difficulty of measuring and modeling systematic errors (e.g., cable attachment position errors) and random errors (e.g., cable sagging) in CDPR, some of which are even unknowable. One promising approach currently being explored is using real data to learn about these errors, as it is an effective way to bridge the gap.

In this paper, we propose a \textbf{f}orward \textbf{k}inematics solving method based on graph neural \textbf{Net}works (GNN) for \textbf{Ca}ble-driven parallel robots, denoted by \textbf{CafkNet}. By utilizing the proposed graph representation for CDPR, CafkNet can rapidly learn its underlying topological structure. Due to the valid training data involved in GNN, the trained CafkNet can efficiently solve the real solutions to the FK problem, rather than finding all solutions. Thanks to the topological similarity of graphs, CafkNet trained from a specific configuration can be easily transferred to other configurations without fine-tuning. Additionally, CafkNet performs well in bridging the Sim2Real gap between vast easily accessible simulation data and limited expensive real-world data. We have conducted extensive simulations and real-world experiments to demonstrate the modeling capabilities of CafkNet. 

\noindent{\textbf{Main contributions}}: 
\begin{itemize}
    \item We propose the CafkNet, a fast and general method to yield real solutions to the FK problem of CDPRs with any configurations via GNN.
    \item We use graph similarity to enable zero/few-shot transfer learning, allowing CafkNet to quickly transfer learned FK knowledge from one configuration to another.
    \item We employ CafkNet to bridge the Sim2Real gap in CDPRs, which has been verified in simulation and real machine experiments.
\end{itemize}

\begin{figure}[!t]
  \centering
  \includegraphics[width=0.485\textwidth]{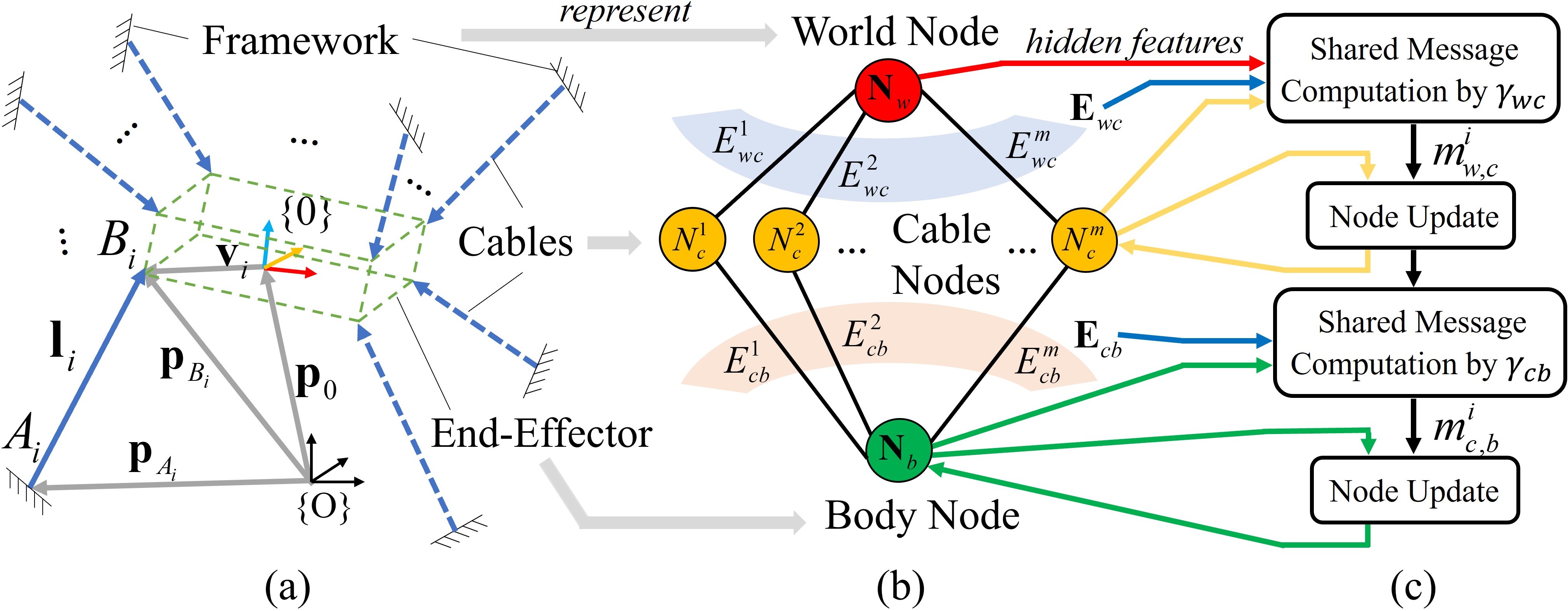}
  \vspace{-20pt}
  \caption{Overview of CafkNet. (a) The geometric model of a spatial CDPR with $m$ cables. (b) The proposed graph presentation for the CDPR (see \prettyref{sec:cdpr_graph}). (c) The top-down computation scheme in each message propagation block. The cable nodes collect data from the world node and are eventually aggregated by the body node (see \prettyref{sec: message propagation}). }
  \label{fig:method}
  \vspace{-18pt}
\end{figure}

The overview of CafkNet is exhibited in \prettyref{fig:method}. 
The rest of the paper is organized as follows. \prettyref{sec:related} reviews related studies about KM of CDPR and GNN applications in robotics. 
% \prettyref{sec:bkg} details the CDPR geometric model. 
\prettyref{sec:method} introduces the proposed graph representation of CDPR and the architecture of CafkNet. Then, \prettyref{sec:exper} demonstrates the experiment details followed by modeling results of CafkNet in  \prettyref{sec:discussion}. Finally, the conclusion and outlook are presented in \prettyref{sec:conclusion}.

%%%%%%%%%%%%%%%%%%%%%%%%%%%%%%%%%%%%%%%%%%%%%%[section]%%%%%%%%%%%%%%%%%%%%%%%%%%%%%%%%%%%%%%%%%%%%%%%%%%%%%%%%%%
\section{Related Work}
\label{sec:related}
\subsection{Kinematic Modeling of CDPR}
KM is a fundamental problem in CDPR's studies and also the first step in each CDPR's application, as revealed in \cite{qian2018review,zhang2022state,zarebidoki2022review}. In CDPR, the IK problem, i.e., the rope length determination given the end-effector pose, is easy to solve, relative to the FK issue. Due to the nonlinearities in FK, there are three potential situations when solving FK \cite{ottaviano2006performance}, i.e., no solution, one solution, and multiple solutions. In addition, \cite{pott2015forward} introduces an FK solver based on energy minimization for a numerical optimization technique. Also, \cite{merlet2023advances} takes into account the nonlinearities of cables and proposes a method for solving the FK of CDPRs with sagging cables. However, these methods are mainly used to model CDPR more finely under their respective configurations, so as to complete the optimization of FK.
More generally, some studies \cite{hong2018review,merlet2023advances} provide the FK solver using graph and graph theory to model the CDPR. Recent works employ diverse machine learning methods trying to solve the system modeling problem, such as neural networks \cite{zare2020kinematic,mishra2022forward}, deep learning \cite{lu2022modlanets}, etc. Although their results validate the feasibility of neural networks on FK, they can not be directly applied to new configurations and need to be retrained when changing robot configurations. In this work, we combine the graph and neural network and utilize GNN to represent and model the CDPR kinematics. Benefitting from the graph similarity, the proposed CafkNet could transfer its learned topological structure of CDPR to other configurations without retraining. This can not be done by NN-based FK solvers.

\subsection{Graph Neural Networks in Robotics}
Recently, data-driven methods have emerged as an effective approach to approximate robotics modeling. Given the fact that the robot’s structure can be easily represented as a graph, where joints are denoted as nodes and links are denoted as edges, GNNs have been increasingly applied in robotics sensing~\cite{yang2023tacgnn,funabashi2022multi}, modeling~\cite{kim2021learning,sanchez2018graph,yang2024one,allen2023graph}, and control~\cite{wang2018nervenet,whitman2023learning,sun2023bridging,ji2021decentralized}. 
~\cite{yang2023tacgnn,funabashi2022multi} propose to use GNN to encode the distributed tactile sensing information for downstream tasks. 
Some works employ GNN to model system dynamics, where graphs represent the rigid body \cite{kim2021learning,sanchez2018graph} and soft cloth \cite{yang2024one}. \cite{allen2023graph} aims to find the representation of robot data, including kinematic data and motion data using the GNN message computation scheme.
NerveNet~\cite{wang2018nervenet} seeks to train a generalized control policy that can transferred to other robot configurations. ~\cite{whitman2023learning,sun2023bridging,ji2021decentralized} propose a similar idea to train a structure-agnostic locomotion policy with a novel message-passing scheme. 
In this paper, we employ the graph to represent the structure of CDPRs and train an FK model based on GNN. Since the training data only contains valid data about cable length and pose, the trained CafkNet can directly find the real solutions to the FK problems rather than struggling in all solutions determination.
To the best of our knowledge, it is \textbf{the first study} to employ the GNN for FK problem of CDPRs.
% This paper is the first work to employ the GNN technique to solve the FK issues for the CDPRs.

%%%%%%%%%%%%%%%%%%%%%%%%%%%%%%%%%%%%%%%%%%%%%%[section]%%%%%%%%%%%%%%%%%%%%%%%%%%%%%%%%%%%%%%%%%%%%%%%%%%%%%%%%%%
\section{Methodology}
\label{sec:method}
\subsection{Geometric Model of CDPR}
Here we consider the tightrope in CDPRs so that the cable model can be represented as the vector.
A 6-degree-of-freedom (DoF) spatial CDPR is illustrated in \prettyref{fig:method}-(a), where two endpoints of the $i$-th cable ($i = 1,2, \cdots, m$) are denoted by $A_i$ and $B_i$. Given the geometric relationship, the cable vector $\mathbf{l}_i$ can be determined as follows.
\begin{align}\label{eq:cable_vector}
    \mathbf{l}_i = \mathbf{p}_{B_i} - \mathbf{p}_{A_i},
\end{align}
where $\mathbf{p}_{B_i} = \mathbf{p}_{0} + R~\mathbf{v}_{i}$. Here $\mathbf{p}_{0} = [x,y,z]^T$ refers to the position of end-effector in the world frame $\{O\}$ (i.e., framework), and $R$ is the rotation matrix from the body frame $\{0\}$ attached on the  end-effector to the world frame $\{O\}$. Additionally, $\mathbf{v}_i$ is a position vector of the vertice $V_i$ expressed in $\{0\}$, which is constant and determined by the cable attachment position on the end-effector. So it is clear that $\mathbf{p}_{A_i}$ and $\mathbf{p}_{B_i}$ are position vectors of $A_i$ and $B_i$ represented in $\{O\}$, respectively.

\subsection{Kinematics of CDPR}\label{sec:kin_cdpr}
Here we can define the pose of end-effector in the task space of CDPR as $\mathbf{Q} = [x,y,z,\phi,\theta, \psi]^T$, where $\phi,\theta,\psi \in[-\pi,\pi]$ are Euler angles of end-effector for roll, pitch, yaw, respectively and they are involved in rotation matrix $R$ in \prettyref{eq:cable_vector}. In addition, its joint space can be defined by a set of cable vectors $\mathbf{L} = \{\mathbf{l}_1, \cdots, \mathbf{l}_m\}$, accordingly. Thus, the IK and FK problems can be formulated as
\begin{align}
    \textbf{IK:~} &\mathbf{L} = f_{IK}(\mathbf{Q}) \label{eq:IKeq}, \\
    \textbf{FK:~} &\mathbf{Q} = f_{FK}(\mathbf{L}) \label{eq:FKeq}.
\end{align}
It is straightforward to find the \prettyref{eq:cable_vector} is equivalent to the function $f_{IK}$ in \prettyref{eq:IKeq}, i.e., solving cable length $\mathbf{l}_i \in \mathbf{L}$ (joint space) from end-effector pose $\mathbf{Q}$ (task space). For FK problem, however, things are no longer so simple. In general, the FK is formulated as an optimization problem, that minimizes the error between the given cable lengths ${\mathbf{L}}$ and lengths determined from IK at one pose $\mathbf{Q}$, that is, 
\begin{align}\label{eq:opti_problem_ik}
    \begin{split}
        \mathop{\arg\min}_{\mathbf{Q}} & \ \| {\mathbf{L}} - f_{IK}(\mathbf{Q}))\|^2, \\
        \textrm{s.t.} \quad & \mathbf{Q}_{lb} \le \mathbf{Q} \le \mathbf{Q}_{ub}.
    \end{split}
\end{align}
Here $\mathbf{Q}_{lb}$ and $\mathbf{Q}_{ub}$ refer to the lower bounds and upper bounds prescribed to the pose $\mathbf{Q}$.

\subsection{Graph Representation of CDPR}
\label{sec:cdpr_graph}
A graph typically consists of nodes and edges. In this work, we define three types of nodes and two types of edges to represent the CDPR as a graph, as outlined in \prettyref{fig:method}-(b). Based on the geometric structure of CDPR in \prettyref{fig:method}-(a), the framework can be considered as a world node $\mathbf{N}_w$, and the end-effector is seen as a body node $\mathbf{N}_b$.  Furthermore, the position of the world frame $\{O\}$ attached to the framework, i.e., $[0,0,0]$, is set as the initial feature of $\mathbf{N}_w$. Similarly, the pose $\mathbf{Q}$ of the body frame attached to the end-effector is selected as the node feature of $\mathbf{N}_b$. For FK modeling, $\mathbf{Q}$ is unknown, so its node feature is initialized to be $[0,0,0,0,0,0]$.
% The predicted pose value is extracted from the body node after message propagation, which will be discussed in \prettyref{sec: message propagation}. 
For CDPRs, cables connecting the framework and the end-effector are defined as a set of cable nodes $\mathbf{N}_c = \{N_{c}^1, N_{c}^2, \cdots, N_{c}^m\}$ for $m$ cables. Here $N_{c}^i$ refers to the $i$-th cable node, and its initial feature is the cable length, i.e., the scalar $||\mathbf{l}_i||$ in \prettyref{eq:cable_vector}, which is known in FK modeling.
In addition, the connection of the cable to framework and the connection to end-effector are considered as two types of edges, $\mathbf{E}_{wc}$ and $\mathbf{E}_{cb}$, whose edge features $e^i_{wc}$ and $e^i_{cb}$ for $i$-th cable are the attachment position vectors $A_i$ and $B_i$ in \prettyref{eq:cable_vector}, respectively. As such, the CDPR can be formulated as a graph $G$ as 
\begin{align}\label{eq:cdpr_graph}
    G=(\mathbf{N}_w,\mathbf{N}_c,\mathbf{N}_b,\mathbf{E}_{wc},\mathbf{E}_{cb}).
\end{align}

\begin{algorithm}[!htbp]
    \caption{GNN-Empowered Forward Kinematic}
    \label{algo:gnn_top_down}
    \begin{algorithmic}[1]
    \State \textbf{Input:} Graph representation of CDPR $G$ in \prettyref{eq:cdpr_graph}
    \State \textbf{Output:} Pose prediction of the end-effector $\mathbf{Q}$    
    \State \textbf{Encoder:}
    \For{each node and edge $u$ in the graph $G$}
        \State Initialize node feature $h_u$ by \prettyref{eq:mlp_input}
    \EndFor    
    \State \textbf{Message Propagation:}
    \For{$s = 1$ to $N$} \Comment{$N$ stacked message propagation blocks}
        \For{each cable node $i$}
            \State Compute message $m_{w,c}^i$ by $\gamma_{wc}$ in \prettyref{eq:m_wc}
            \State Update cable node feature $h_c^i$ by \prettyref{eq:h_c}
        \EndFor
        \State Initialize an empty set $\{\}$ for body node messages
        \For{each cable node $i$}
            \State Compute message $m_{c,b}^i$ by $\gamma_{cb}$ in \prettyref{eq:m_cb}
            \State Add $m_{c,b}^i$ to the set $\{m^i_{c,b}\}$
        \EndFor
        \State Update body node feature $h_b$ by \prettyref{eq:h_b}
        \EndFor        
    \State \textbf{Decoder:}
    \State Decode the final body node feature to pose prediction of the end-effector by \prettyref{eq:mlp_output}.    
    \end{algorithmic}    
\end{algorithm}

\subsection{GNN-Empowered Forward Kinematic Modeling}
\label{sec: message propagation}
In this subsection, we will parameterize the FK problem of CDPRs as a GNN using the Encoder-Propagation-Decoder format \cite{battaglia2018relational} for message passing, as summarized in \prettyref{algo:gnn_top_down}.

\textbf{Encoder}: The encoder receives a graph $G$ in \prettyref{eq:cdpr_graph} representing a CDPR, then it maps each node feature and edge feature to a fixed-dimension embedding by

\begin{equation}\label{eq:mlp_input}
    h_u = \sigma_{in}(x_u), ~u\in \{w, c, b, wc, cb\},
\end{equation}
where $\sigma_{in}$ denotes the initialization multi-layer perceptron (MLP). Here $x_u$ and $h_u$  refer to the initial and hidden features of node (or edge) $u$, respectively.

\textbf{Message Propagation}: The propagation model consists of $N$ stacked message propagation blocks to update node embedding, which means message computation and aggregation are recurrently applied $N$ times. In each message propagation block, a top-down message-passing scheme is adopted, as exhibited in \prettyref{fig:method}-(c).
% A Top-Down message-passing scheme is adopted for the FK problem, which means the cable nodes collect data from the world node and are eventually aggregated by the body node. We use two MLPs to parameterize the relationship between nodes and the same type of edge shares parameters. 
% The message propagation block is shown in Fig.~\ref{fig:method}-(c).
%
Specifically, the model first computes the message from the world node to each cable node $i$ by
\begin{equation}\label{eq:m_wc}
    m^i_{w,c} = \gamma_{wc}(h_w,h^i_c,e^i_{wc}),
\end{equation}
where $\gamma_{wc}$ refers to the MLP from world to cable nodes. Then, $h_w$ and $h^i_c$ are hidden features of the world node and the $i$-th cable node, respectively, then $e^i_{wc}$ is the edge feature of the edge between the world node and the $i$-th cable node. 
Once every node finishes computing messages, the hidden feature of every cable node $h^i_c$ can be updated based on the message computation and itself:
\begin{equation}\label{eq:h_c}
    h^i_c = \max(\{h^i_c, m^i_{w,c}\}),
\end{equation}
% where $h^i_c$ denotes the hidden feature of cable node $i$, $m_{w,c}$ denotes the propagated message from world node. 
Similarly, the message from the $i$-th cable node to the body node is computed by
\begin{equation}\label{eq:m_cb}
    m^i_{c,b} = \gamma_{cb}(h^i_c,h_b,e^i_{cb}), 
\end{equation}
where $\gamma_{cb}$ denotes the MLP from cable to body nodes, and $h_b$ are hidden features of the body node. Then $e^i_{cb}$ refers to the edge feature of edge from $i$-th cable node to body node.
After messages of all cable nodes $\{m^i_{c,b}\}$ are calculated, the body node is updated via
\begin{equation}\label{eq:h_b}
    h_b = \max(\{h_b,\{m^i_{c,b}\}\}).
\end{equation}
% where the $\{m^i_{c,b}\}$ contains messages from all cable nodes.
Note that, as shown from line 9 to line 18 in \prettyref{algo:gnn_top_down}, we only employ two message computation functions, i.e., $\gamma_{wc}$ and $\gamma_{cb}$, to handle the message propagation on two types of edges in the proposed graph representation. Thus, the same type of edges share MLP parameters.

% All edges of the same edge type share the same instance of the message computation function, i.e. the trainable parameters of MLP. We only have two MLP instances for this graph: $\gamma_{wc}$ and $\gamma_{cb}$.

\textbf{Decoder}: 
In the decoder, the output MLP $\sigma_{out}$ takes the final hidden feature of body node $h_b$ after $N$ propagation blocks as input and outputs the pose of end-effector $\mathbf{Q}$, as formulated by
\begin{equation}\label{eq:mlp_output}
    \mathbf{Q} = \sigma_{out}(h_b).
\end{equation}

% \subsection{Training Details}
\subsection{Transfer Learning across Configurations}
CafkNet employs parameter sharing among identical node relationships, enhancing its ability to perform structural transfer tasks effectively. Specifically, there are only two parameterized MLPs in our graph for message propagation, including $\gamma_{wc}$ for the relationship between world node and cable node, as well as $\gamma_{cb}$ for the relationship between cable node and body node. The parameters of MLPs are independent of cable number. As a result, our approach facilitates the application of the kinematic model to various CDPR configurations, significantly improving generalization capabilities across different cable arrangements. Extensive experiments in \prettyref{sec:discussion} verify the transferability of our CafkNet.

%%%%%%%%%%%%%%%%%%%%%%%%%%%%%%%%%%%%%%%%%%%%%%[section]%%%%%%%%%%%%%%%%%%%%%%%%%%%%%%%%%%%%%%%%%%%%%%%%%%%%%%%%%%
\section{Experiments}
\label{sec:exper}

\subsection{Dataset}
\label{sec:dataset}
\begin{figure}[!tbp]
    \centering
    % \vspace{-5pt}
    \includegraphics[width=0.45\textwidth]{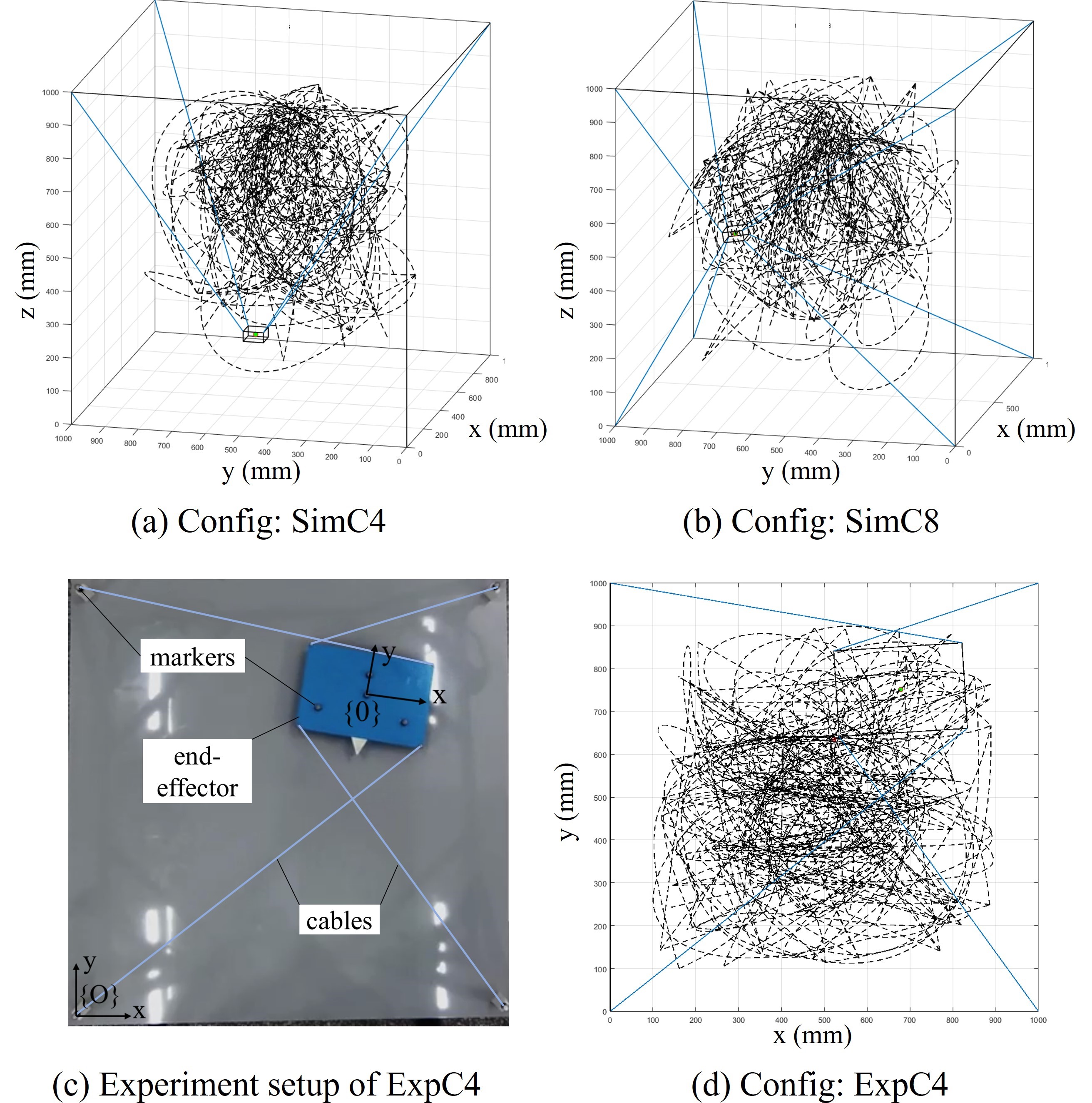}
    \vspace{-10pt}
    \caption{Configurations of CDPR in simulation and experiment scenarios. In simulation cases, we randomly generate $100$ polynomial trajectories (dashed curves) and sample $100$ poses (including position and orientation) along each trajectory.}
    \label{fig:dataset}
    \vspace{-15pt}
\end{figure}

In this work, we consider $8$ configurations of CDPR, including spatial ones (e.g., \prettyref{fig:dataset}-(a) and (b)) and planar ones (e.g., \prettyref{fig:dataset}-(c) and (d)). We investigate under-constrained, fully-constrained and over-constrained cases, ranging from four cables to ten cables. 
% The cable attachment locations in each configuration are detailed in \prettyref{app:config_CDPRs}. 
In addition, we validate our method in the real experiment setup, as shown in \prettyref{fig:dataset}-(c)

In the simulator, we randomly generate $100$ polynomial trajectories for each configuration, as shown in dashed curves in \prettyref{fig:dataset} and then uniformly sample $100$ poses along each trajectory. Furthermore, the cable lengths at each pose can be simply calculated via IK at each configuration. In the real experimental setup, the pose of the end-effector could be determined by the external motion capture system, OptiTrack (NaturalPoint Inc., USA), using attached markers in the end-effector and anchors, as depicted in \prettyref{fig:dataset}-(c). The cable lengths are measured by the motor encoders accordingly.
We partition the dataset into a training set comprising $80\%$ of the data and a testing set containing the remaining samples.

\subsection{Metrics}
In this work, we use a supervised learning paradigm to train CafkNet for CDPRs based on the above dataset. Here, the Root Mean Square Error (RMSE) is selected as our performance metric. Note that we only calculate the RMSE loss about positions, disregarding any variations in orientations.

\subsection{Implementation Details}
In our implementation, we utilize Adam \cite{kingma2014adam} is utilized as the optimizer, with a batch size of $32$. Other parameters, such as the hidden feature dimensions and number of MLP layers in CafkNet, the initial learning rate, and the reduction factor for the learning rate scheduler, are optimized automatically via the toolkit in \cite{liaw2018tune}.
All codes are developed using PyTorch and trained on a single NVIDIA GeForce RTX 4090 GPU card.

%%%%%%%%%%%%%%%%%%%%%%%%%%%%%%%%%%%%%%%%%%%%%%[section]%%%%%%%%%%%%%%%%%%%%%%%%%%%%%%%%%%%%%%%%%%%%%%%%%%%%%%%%%%
\section{Results and Discussion}
\label{sec:discussion}

\subsection{FK Modeling for CDPR}
% In this subsection, we compare CafkNet with two other FK solvers based on the MLP modeling and the optimization-based method, respectively. 
% % for solving the kinematic modeling problem of CDPRs. We conduct experiments using CDPRs with 6, 7, and 8 cables as examples. 
% The resulting modeling accuracy and time cost are summarized in \prettyref{tab:results}.

In this subsection, we present the results and time cost of solving the FK problem at numerous CDPR configurations by CafkNet. Additionally, we compare it with the MLP method and the optimization-based FK solver separately. 

For CafkNet, we provide two types of FK solutions. The first approach involves training CafkNet using data from a single configuration and solving the FK problem specifically for that configuration. Therefore, each model is dedicated to solving the FK problem for a single configuration, and this approach is referred to as CafkNet (one2one). The second approach involves training a universal CafkNet model using data from all configurations. During the testing phase, the model is divided into multiple heads, each addressing the FK issues for a specific configuration. This method is referred to as CafkNet (multi-task).

As a comparison, we also employ MLP to solve FK problems, where all node features and edge features are as input and output is the position of the end-effector. In addition, an optimization-based FK solver is also used, which takes advantage of the least square method in \cite{virtanen2020scipy} to solve the optimization problem as formulated in \prettyref{eq:opti_problem_ik}, where the  $\mathbf{Q}_{lb}$ and $\mathbf{Q}_{ub}$ are set as $[100,100,100,-0.26,-0.26,0]$ and $[900,900,900,0.26,0.26,0]$, respectively. The initial guess is given as $[500,500,500,0,0,0]$.

The results of modeling accuracy, i.e., the RMSE loss of predicted positions and reference positions, as revealed in \prettyref{tab:results}. Here we can observe that the opt. FK solver achieves the smallest error in SimC6 to SimC10 but fails in SimC4 and SimC5, where CafkNet achieves the lowest error. But from SimC6 to SimC10, the errors of CafkNet remain within $1.5$ mm. On the other hand, the MLP solver exhibits the poorest accuracy, with errors exceeding $15$ mm. Furthermore, we notice that the accuracy of CafkNet (multi-task) is comparable to CafkNet (one2one). 

However, considering the computational time for FK solving, as shown in \prettyref{tab:time_cost}, the opt. FK solver takes an average time of approximately $1$ s for the best-accuracy configurations, SimC4 $\sim$ SimC10, but experiences a significant time raising at failure cases. In contrast, it can be found that CafkNet (one2one) achieves remarkably low computation time, with only $0.649$ s for SimC4 and a few milliseconds for other configurations. Considering both computational time and solution accuracy, the CafkNet offers a trade-off between low time consumption and relatively high precision, demonstrating significant potential for real-time CDPR applications compared to the opt. FK solver.

\begin{figure*}
    \centering
    % \vspace{-5pt}
    \includegraphics[width=0.96\textwidth]{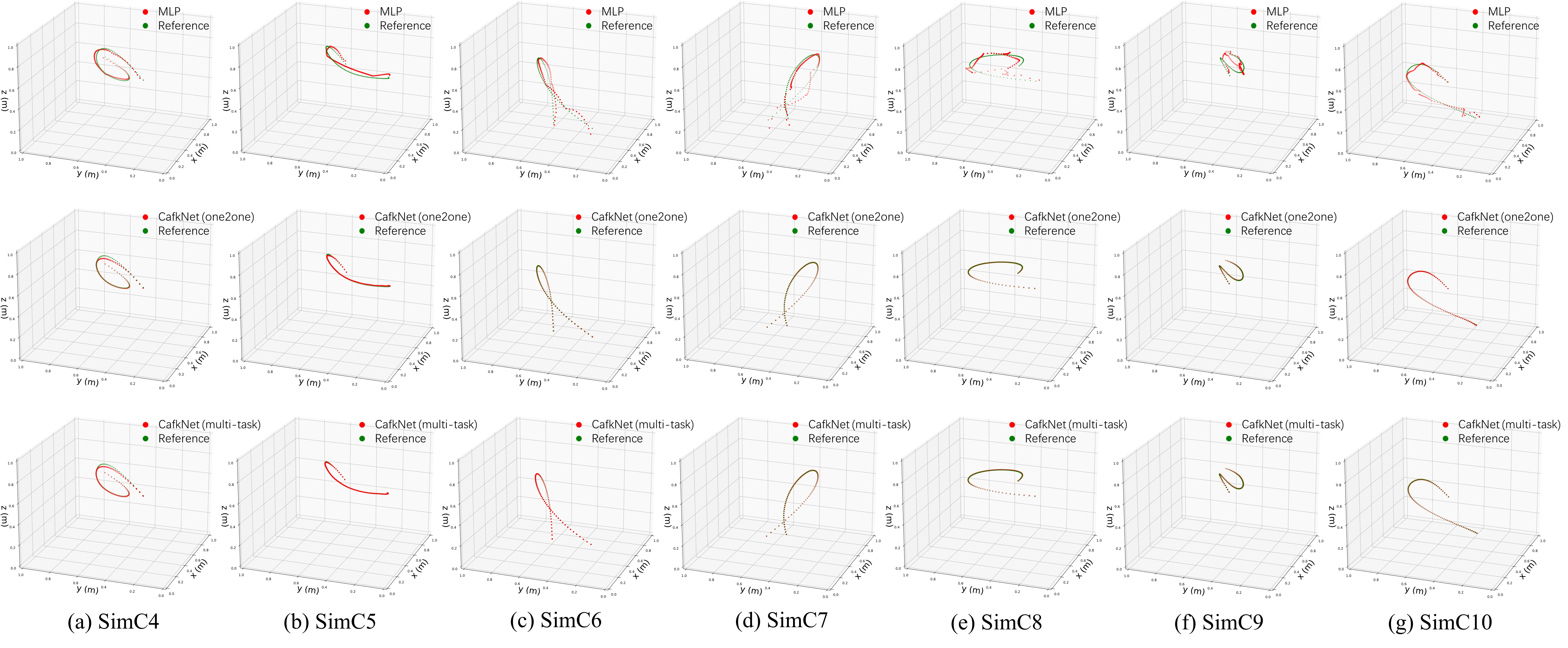}
    \vspace{-10pt}
    \caption{Reference trajectories and resulting trajectories solved by MLP and CafkNet under configurations (a) - (g). Top row: by MLP; Middle row: by CafkNet (one2one); Bottom row: by CafkNet (multi-task).}
    \label{fig:trajectory}
    \vspace{-15pt}
\end{figure*}

Note that the \prettyref{tab:time_cost} indicates the overall time taken by both methods to compute a complete trajectory, comprising 100 FK problems. The resulting trajectories are visually presented in \prettyref{fig:trajectory}, illustrating FK solutions solved by MLP, CafkNet (one2one), and CafkNet (multi-task), respectively. The close correspondence observed between the predicted and reference trajectories serves as evidence of the accurate FK solutions.

\begin{table}[!htbp]
    \caption{Modeling Accuracy (unit: mm). \\Bold: minimum error in each configuration.}
    \vspace{-5pt}
    \centering    
    \resizebox{0.44\textwidth}{!}{%
    \begin{tabular}{@{}ccccccccc@{}}
    \toprule
        \multicolumn{2}{c}{\multirow{2.5}{*}{Method}} & \multicolumn{7}{c}{Configuration} \\ \cmidrule{3-9}
        &~ & \multicolumn{1}{c|}{SimC4} & \multicolumn{1}{c|}{SimC5} &  \multicolumn{1}{c|}{SimC6}  & \multicolumn{1}{c|}{SimC7} & \multicolumn{1}{c|}{SimC8} & \multicolumn{1}{c|}{SimC9} & \multicolumn{1}{c}{SimC10} \\ 
        \midrule\midrule
        \multicolumn{2}{c}{Opt. FK Solver} & 443.00 & 349.16 & \textbf{0.07} & \textbf{0.00} & \textbf{0.00} & \textbf{0.00} & \textbf{0.00}  \\ \cmidrule{1-9}
        \multicolumn{2}{c}{MLP} & 18.67 & 16.45 & 11.24 & 17.14 & 31.51 & 31.49 & 15.37  \\ \cmidrule{1-9}
        \multirow{2.5}{*}{CafkNet} & \multicolumn{1}{c}{one2one} & \textbf{4.19} & 2.46 & {0.42} & 1.57 & {1.58} & {1.61} & 1.05 \\ \cmidrule{2-9}
        ~ & \multicolumn{1}{c}{multi-task} & 4.97 & \textbf{1.27} & 0.86 & {1.45} & 2.39 & 2.32 & {0.55} \\    
    \bottomrule
    \end{tabular}
    }
    \label{tab:results}
    \vspace{-10pt}
\end{table}

\begin{table}[!ht]
    \caption{Time Cost of FK solving (unit: s).}
    % \vspace{-5pt}
    \centering    
    \resizebox{0.44\textwidth}{!}{%
    \begin{tabular}{@{}cccccccc@{}}
    \toprule
        \multicolumn{1}{c}{\multirow{2.5}{*}{Method}} & \multicolumn{7}{c}{Configuration} \\ \cmidrule{2-8}
        & \multicolumn{1}{c|}{SimC4} & \multicolumn{1}{c|}{SimC5} &  \multicolumn{1}{c|}{SimC6}  & \multicolumn{1}{c|}{SimC7} & \multicolumn{1}{c|}{SimC8} & \multicolumn{1}{c|}{SimC9} & \multicolumn{1}{c}{SimC10} \\ 
        \midrule\midrule
        \multicolumn{1}{c}{Opt. FK Solver} & 51.932 & 76.245 & 1.171  & 1.087  & 0.993  & 0.999  & 1.084   \\ \cmidrule{1-8}
        \multicolumn{1}{c}{CafkNet (one2one)} & 0.649 & 0.002 & 0.004  &  0.001 & 0.001  & 0.002  & 0.004  \\ 
    \bottomrule
    \end{tabular}
    }
    \label{tab:time_cost}
    % \vspace{-15pt}
\end{table}

\subsection{Transfer Learning in CDPR}
% 1. sim - sim+noise (table)
% 2. sim - real data (fig:exp setup + sim config)
% \subsubsection{Dataset}
% \subsubsection{Results}
% \subsubsection{Discussion}
In this subsection, we investigate the transferability of our CafkNet, and the resulting modeling errors are given in \prettyref{tab:transfer}. In the GNN, all cable and framework node interactions share parameters $\sigma_C$, which allows us to transfer the kinematic model to CDPRs with different cable configurations to assess the generalization ability of the trained model. We conduct a series of experiments on transferability tests, where the kinematic model is transferred without further fine-tuning.

We conduct two groups of zero-shot transfer experiments. In the first group (Exp C-1), we train CafkNet on one configuration, e.g., SimC$i$, and then directly apply it to solve the FK problem on other configurations SimC$j$, where $j \neq i$. We calculate the average errors for the resulting positions, i.e., $x, y$, and $z$. In the second group of experiments (Exp C-2), we aim to predict the pose in the configuration SimC$i$ using data from multiple configurations SimC$j$, where $j \in (1,2, ..)$ and $j \neq i$ for training. The experiment results are revealed in \prettyref{tab:transfer}, where rows 2 to 8 display outcomes of Exp C-1, and row 9 displays the results of Exp C-2.

For example, the second row in \prettyref{tab:transfer} represents the training of CafkNet on the data from SimC4, followed by directly solving the FK problems on SimC5-SimC10, respectively. It can be observed that the errors in Exp C-1 are relatively high. This result aligns with our expectations, as CafkNets have limited and singular learning data on a single configuration, resulting in better performance on the same configuration and poorer performance on other configurations.

However, from the results of Exp C-2, i.e., last row in \prettyref{tab:transfer}, we discover that due to the increased variety and volume of training data in SimC$j$ for CafkNet, even without prior exposure to configuration SimC$i$, it can still successfully transfer and solve the FK problem in the unseen configuration, showcasing good zero-shot transferability.

It is worth noting that despite the less satisfactory results in Exp C-1, we can also observe that CafkNet, aside from performing well on its own configuration, exhibits increasing accuracy errors as the distance from its own configuration increases. For instance, the model trained on SimC6 demonstrates better accuracy on SimC7 compared to SimC8, and the results on SimC5 are superior to those on SimC4.

\begin{table}[!ht]
    \caption{Modeling Accuracy on Transferability of CafkNet (unit: mm). Bold: minimum test error in each configuration.}
    % \vspace{-5pt}
    \centering    
    \resizebox{0.44\textwidth}{!}{%
    \begin{tabular}{@{}cccccccc@{}}
    \toprule
        \diagbox[width=17mm]{Train}{Test}& \multicolumn{1}{|c|}{SimC4} & \multicolumn{1}{c|}{SimC5} & \multicolumn{1}{c|}{SimC6} &  \multicolumn{1}{c|}{SimC7}  & \multicolumn{1}{c|}{SimC8} & \multicolumn{1}{c|}{SimC9} & \multicolumn{1}{c}{SimC10} \\ \midrule\midrule
        \multicolumn{1}{c|}{SimC4} & - & 14.85 & 36.11 & 46.37 & 41.82 & 53.24 & 44.53  \\ \cmidrule{1-8}
        \multicolumn{1}{c|}{SimC5} & 20.64 & - & 39.03 & 49.7 & 36.23 & 60.19 & 50.97  \\ \cmidrule{1-8}
        \multicolumn{1}{c|}{SimC6} & 64.94 & 58.45 & - & 21.07 & 31.77 & 25.24 & 22.72  \\ \cmidrule{1-8}
        \multicolumn{1}{c|}{SimC7} & 80.06 & 81.74 & 14.95 & - & 43.61 & 9.29 & 16.45  \\  \cmidrule{1-8}
        \multicolumn{1}{c|}{SimC8} & 89.99 & 88.05 & 47.17 & 33.76 & - & 34.09 & 21.91  \\  \cmidrule{1-8}
        \multicolumn{1}{c|}{SimC9} & 20.98 & 67.05 & 7.13 & 9.46 & 23.38 & - & 6.04  \\ \cmidrule{1-8}
        \multicolumn{1}{c|}{SimC10} & 12.29 & 97.71 & 5.58 & 8.12 & 13.17 & 7.65 & -  \\ \cmidrule{1-8}
        \multicolumn{1}{c|}{multi-config.} & \textbf{4.14} & \textbf{8.23} & \textbf{2.06} & \textbf{1.74} & \textbf{11.96} & \textbf{1.09} & \textbf{0.69} \\
    \bottomrule
    \end{tabular}
    }
    \label{tab:transfer}
\end{table}

\subsection{Bridging Sim2Real Gap}
% We have two datasets w/ noise
In practical applications of CDPR, noise is inevitable. It can originate from various sources such as the mechanical structure, including the winch and pulley, as well as from the motor encoder and cable extension. Thus the sim2real gap in CDPR should be well studied. To validate the robustness of CafkNet against noisy data, we utilize two datasets in this study. One dataset was generated by injecting Gaussian noise into the data obtained from the simulator. The other dataset was collected from a real experimental platform, as depicted in \prettyref{fig:dataset}-(c).

% 1: Injected Noise in simulated data
\subsubsection{Simulated Data with Artificial Noises}
In \prettyref{fig:dataset}-(c), we find that the pose of CDPR can be accurately captured by the motion capture system, so we do not add noise to the pose in the simulated data.
On the contrary, the main errors in the real experiment come from the length of the cable, including the error of the motor encoder, as well as the position of the cable on the pulley and where it is mounted at the end-effector. Thus, we introduce the random error $\epsilon$ in the simulator mimicking cable attachment position noises, i.e., $A_i \pm \epsilon,~B_i \pm \epsilon$, where $\epsilon \sim \mathcal{N}(\mu = 0, \sigma^2)$ and $\mathcal{N}$ refers to the Gaussian distribution.
We conduct two sets of experiments targeting different $\sigma$ values to assess the performance of CafkNet when exposed to artificial noise, i.e., $\sigma = 5$ and $\sigma = 10$, respectively. In each experimental set, we also consider three categories based on the source of the training and testing data for each CDPR configuration. These categories include:
\begin{itemize}
    \item Training and testing on simulated data.
    \item Training on simulated data and testing on noise data.
    \item Training and testing on noise data.
\end{itemize}
Here, we define the data collected from the simulator, i.e., from \prettyref{fig:dataset}-(d), as simulated data, while the data with artificially added random noise is referred to as noise data.

% Here, $\mu = 0, \sigma = 5$. In addition, we conducted three experiments for each configuration as follows, i.e., Case 1: training on clean data and testing on clean data; Case 2: training on clean data and transferring to noise data; Case 3: training on noise data and testing on noise data.

\begin{figure}[!tb]
    \centering
    % \vspace{-5pt}
    \includegraphics[width=0.44\textwidth]{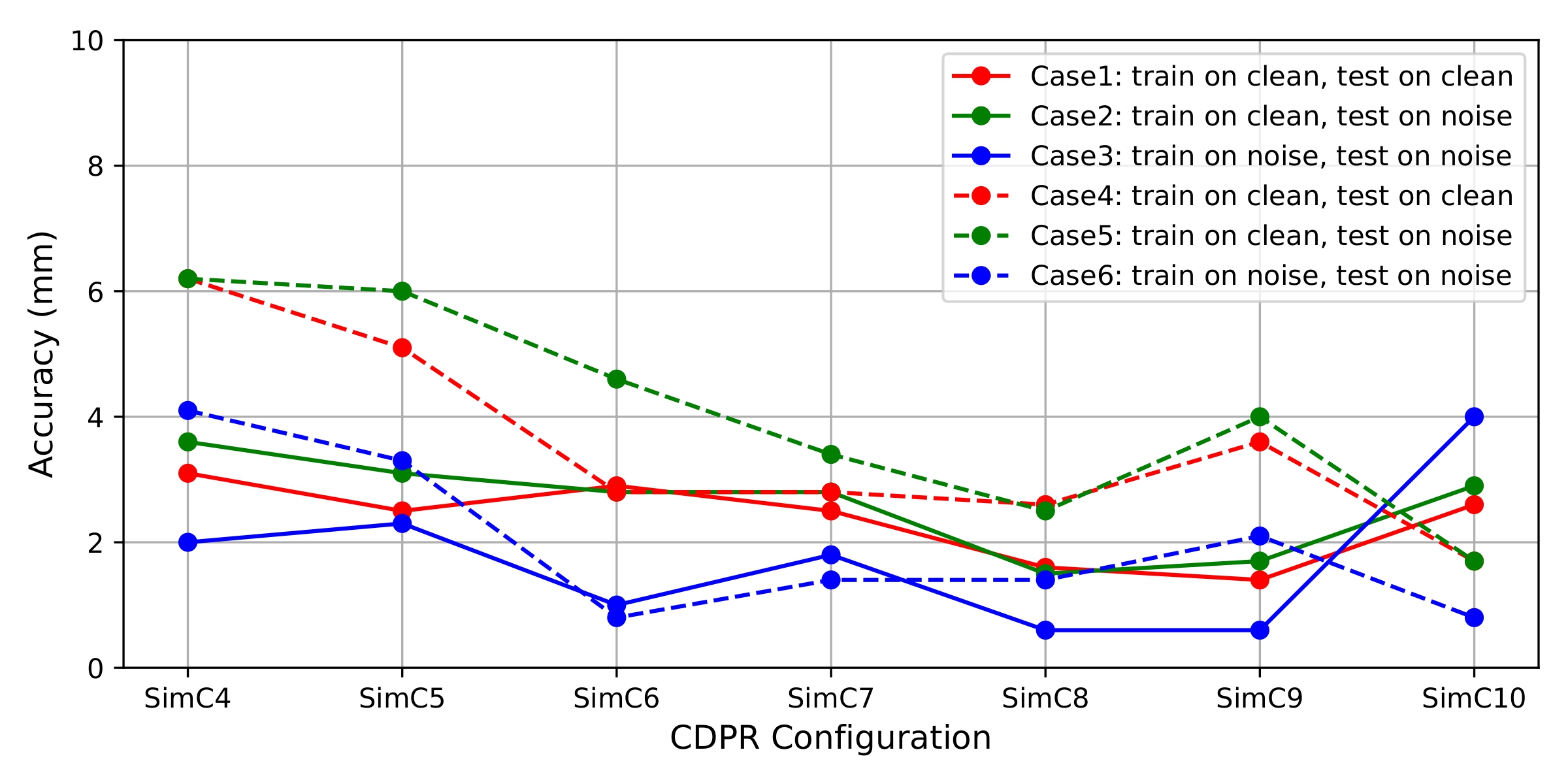}
    \vspace{-10pt}
    \caption{Modeling accuracy of CafkNet working on artificial noise data. Here errors in case 1 to case 3 are generated by $\epsilon \sim \mathcal{N}(0, 5^2)$, and errors in case 4 to case 6 are generated by $\epsilon \sim \mathcal{N}(0, 10^2)$. $\mathcal{N}$ refers to the Gaussian distribution.}
    \label{fig:artificial_noise}
    \vspace{-15pt}
\end{figure}

The results are demonstrated in \prettyref{fig:artificial_noise}. 
From it, we can observe that as the magnitude of artificially introduced noise increases, the modeling error solved by CafkNet also exhibits a corresponding overall increase. However, the overall increment in modeling error does not exceed a relationship greater than twice the magnitude of the artificial error (i.e., $5$ to $10$), particularly when observing the green curve.
In addition, as the number of cables increases, the modeling error from CafkNet gradually decreases and stabilizes below 4mm. 
We can also find that the resulting curves of case 1 and case 2 are very close, with almost identical computational errors. This indicates that the model trained on clean data demonstrates generalization capability, allowing it to generalize to data with errors even in zero-shot scenarios. A similar phenomenon can be found between case 3 and case 4, as well. As anticipated, if noise data is used during training, CafkNet will provide the lowest error results, as observed in the blue curve.

However, noise data in real-world applications is often obtained from real machine experiments, which can be costly. On the other hand, simulated data is relatively easier to acquire, and its quantity can surpass that of real machine data. Therefore, in the next, we will demonstrate how CafkNet combines low-cost simulated data with a portion of machine data to achieve accurate FK solutions for real-world scenarios.

\begin{figure}[!tb]
    \centering
    % \vspace{-5pt}
    \includegraphics[width=0.48\textwidth]{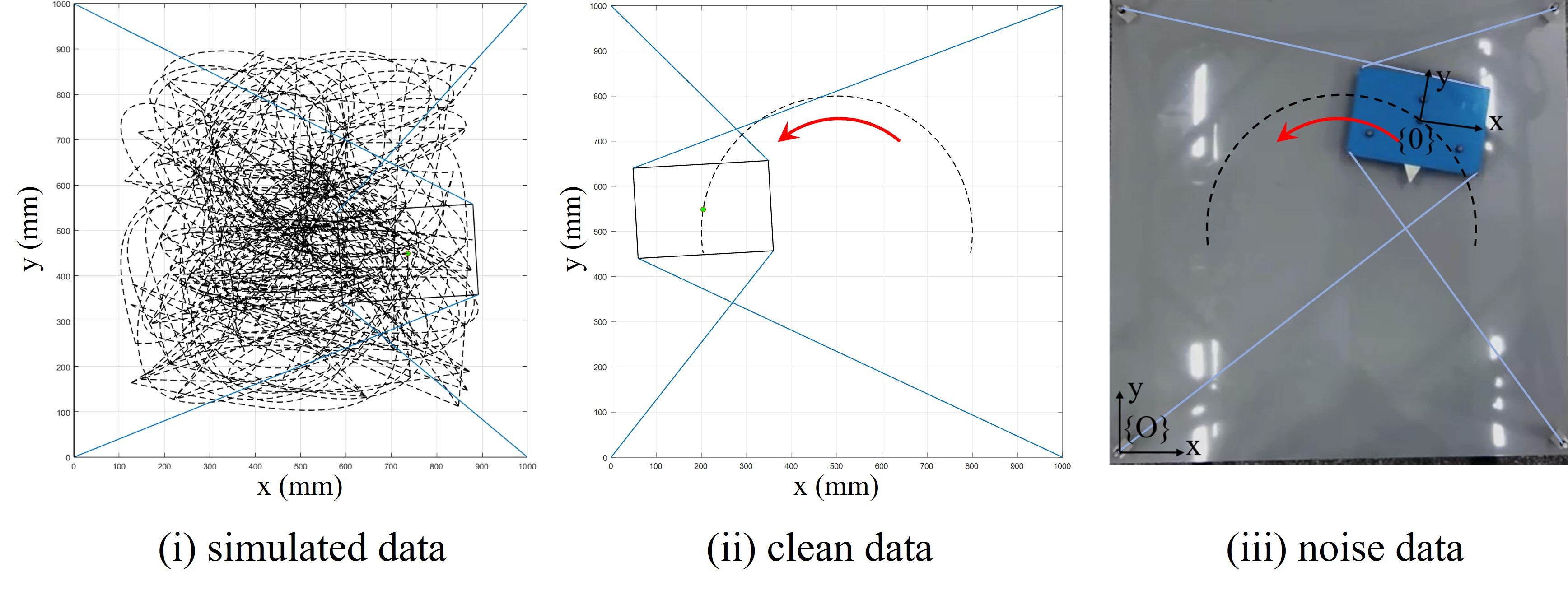}
    \vspace{-20pt}
    \caption{Data sources of Sim2Real experiment for the configuration of ExpC4. (i) Simulated data. In the simulator, there are $100$ random trajectories and $100$ samplings per trajectory. (ii) Clean data. In the simulator, we have devised a target trajectory (dashed curve) to control the real CDPR to execute it. Here $1000$ samplings are generated. (iii) Noise data. In the real setup, the end-effector of CDPR follows the target trajectory. During the experiment, a total of 1103 data are collected. Above all, each data composes the end-effector pose ($x,y,\theta$) and the corresponding length of $4$ cables.}
    \label{fig:three_cases_sim2real}
    \vspace{-15pt}
\end{figure}

% 2: experimental data
\subsubsection{Real Experimental Data}
In this subsubsection, we will present the modeling process for deploying a real CDPR platform, which primarily involves building a model in the simulator based on design blueprints, i.e., \prettyref{fig:three_cases_sim2real}-(i), generating simulated trajectories based on tasks in the simulation environment, and conducting simulated operations (\prettyref{fig:three_cases_sim2real}-(ii)). Subsequently, the target trajectories are executed on the actual machine (\prettyref{fig:three_cases_sim2real}-(iii)). During this process, the CDPR model in the simulation environment can be a replica of the model in the design blueprints, but there will inevitably be parameter discrepancies between the simulated environment and the real platform. Therefore, we demonstrate the sim2real capability of CafkNet.

For this experiment, we primarily have three sources of data, as depicted in \prettyref{fig:three_cases_sim2real}. One type of data is defined as simulated data, which consists of randomly sampled trajectories (i.e., pose-length pairs) generated on the simulator, as shown in \prettyref{fig:three_cases_sim2real}-(i). Another type of data is defined as clean data, which is generated on the simulation platform specifically for the target trajectory, as illustrated in \prettyref{fig:three_cases_sim2real}-(ii). These first two types of data are readily available and do not consider any errors. The last type of data, referred to as noise data, as depicted in \prettyref{fig:three_cases_sim2real}-(ii), originates from measurements when the practical CDPR executes the target path. This data includes both systematic errors inherent in the machine (e.g., installation errors in pulleys, shape errors in end-effector) and random errors during execution (e.g., motor encoder errors, measurement errors in the motion compensation system). Systematic errors are unique to each device, and their distribution is specific to the current equipment. In other words, a different set of systematic errors would be observed in another device. Moreover, it is challenging to model random errors.

To evaluate the sim2real capability of CafkNet, we design three methods based on the training data, as summarized in \prettyref{tab:transfer_method_result}. Then, in the testing, cable lengths from either the clean data or noise data are inputted into the trained model. Then it solves for the end-effector poses and compares them with the corresponding poses from the clean data and noise data to compute the average error, respectively. Three methods are described in detail below.
\begin{itemize}
    \item $sim2real$: During the training phase, CafkNet is trained using all simulated data and tested all clean data.
    \item $real2real$: In the training phase, $80\%$ of the clean data is used, while in the testing phase, the remaining $20\%$ of the clean data or the noise data is used.
    \item $sim\&real2real$: This method introduces training data including all simulated data and $80\%$ of clean data. The testing data remains consistent with the $real2real$ method.
\end{itemize}

\begin{figure}[!tbp]
    \centering
    % \vspace{-5pt}
    \includegraphics[width=0.46\textwidth]{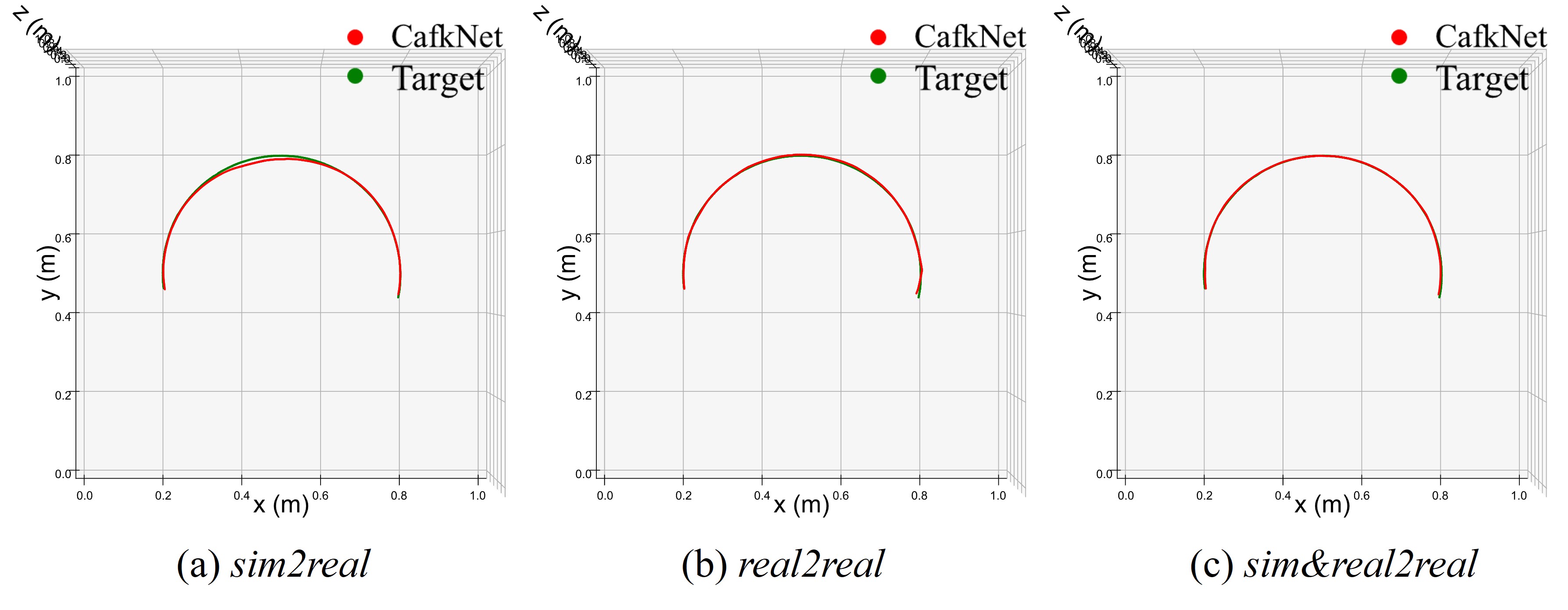}
    \vspace{-10pt}
    \caption{Target trajectory and resulting trajectory solved by CafkNet under three methods.}
    \label{fig:half_circle_three_methods}
    \vspace{-15pt}
\end{figure}

The resulting trajectories solved by CafkNet under these three methods are presented in \prettyref{fig:half_circle_three_methods}, and their accuracy results are given in \prettyref{tab:transfer_method_result}. The experimental findings demonstrate that CafkNet, trained using composite data consisting of simulated data and $80\%$ of the clean data, yields end-effector poses that closely align with real machine data. This indicates its effectiveness in bridging the sim2real gap in practical applications of CDPRs. 
Specifically,  when testing on clean data, as expected, the ${sim2real}$ method exhibits the lowest error, followed by ${sim\&real2real}$, while ${real2real}$ performs the poorest. This can be attributed to the fact that the quantity of simulated data ($10000$) is ten times that of the clean data ($1000$). Consequently, the ${sim2real}$ method has more information about end-effector poses and cable lengths, which contributes to its superior performance on clean data (error in $1.7$ mm). Compared to the ${real2real}$ method, the ${sim\&real2real}$ method supplements the data, resulting in a decrease in modeling error from $2.7$ to $2.4$ mm.

When directly transferring the model to unseen noise data, the method ${sim\&real2real}$  achieves the best performance, with an accuracy error as low as $1.3$ mm. This can be attributed to the method's rich dataset, which not only includes a large range of motion information but also incorporates specific information about the target trajectory. These factors contribute to the model's ability to minimize FK-solving errors during sim2real transfer.

Through this experiment, we have demonstrated that our CafkNet can effectively bridge the sim2real gap by leveraging inexpensive and readily available simulated data combined with specific data. It can compensate for certain errors in CDPRs with given configurations. This opens up potential applications for CafkNet in various domains.

\begin{table}
    \caption{Sim2Real Experiment Details and Results. Cases \MakeLowercase{i, ii, iii} are revealed in \prettyref{fig:three_cases_sim2real}. Bold: minimum error in each testing data.}
    % \vspace{-5pt}
    \centering    
    \resizebox{0.44\textwidth}{!}{%
    \begin{tabular}{@{}cccccc@{}} % 10 columns
    \toprule
        \multirow{2.5}{*}{Methods} & \multirow{2.5}{*}{Train on} & \multicolumn{2}{c|}{Test on} & \multicolumn{2}{c}{Results (unit: mm)}\\\cmidrule{3-4} \cmidrule{5-6} 
         &  &  Clean Data & \multicolumn{1}{c|}{Noise Data} &  Clean Data & Noise Data \\ 
        \midrule\midrule
        ${sim2real}$ & i  & ii  &  \multicolumn{1}{c|}{iii} & \textbf{1.7} & 1.6 \\ \cmidrule{1-6} 
        ${real2real}$ & $80\%$ ii  & $20\%$ ii & \multicolumn{1}{c|}{iii} & 2.7 & 2.2 \\ \cmidrule{1-6} 
        $sim\&real2real$ & $\text{i} + 80\%$ ii  & $20\%$ ii  & \multicolumn{1}{c|}{iii} & 2.4 & \textbf{1.3} \\
    \bottomrule
    \end{tabular}
    }
    \label{tab:transfer_method_result}
    % \vspace{-15pt}
\end{table}

%%%%%%%%%%%%%%%%%%%%%%%%%%%%%%%%%%%%%%%%%%%%%%[section]%%%%%%%%%%%%%%%%%%%%%%%%%%%%%%%%%%%%%%%%%%%%%%%%%%%%%%%%%%
\section{Conclusion}
\label{sec:conclusion}
% mention: IK does not work, b/z 1. IK of CDPR is simple, 2. graph of IK ...

This paper presents the CafkNet, a GNN-empowered FK-solving method for CDPRs with superior generality, high
accuracy, and low time cost. It is the first work to utilize GNN to handle the FK issue of CDPRs. The study begins by introducing a graph structure that represents the topology of CDPRs, followed by network learning on the graph via message propagation block. The experimental results demonstrate that CafkNet can learn the inherent topological relationships of CDPRs and efficiently find real solutions to the FK problem. 
The results are validated across different types of CDPRs, including spatial CDPRs and planar CDPRs, as well as under configurations such as under-constrained, fully-constrained, and over-constrained CDPRs. The validation is conducted in both simulation environments and real-world scenarios. This work only considers the straight cables, so the appropriate graph representation for sagging cables in FK problem may be a future work.

% This work focuses solely on the FK problem, which is known to be challenging in the kinematic modeling of CDPRs. So, in the future, we aim to apply the GNN to solve the IK problem and dynamic modeling issues.

\bibliographystyle{IEEEtran}
\typeout{}
\bibliography{IEEEabrv,mybibfiles}
\end{document}